\documentclass[journal]{IEEEtran}
\ifCLASSINFOpdf
\else
\fi
\usepackage{url}

\usepackage{epsfig}
\usepackage{graphicx}
\usepackage{amsmath}
\usepackage{amssymb}

\usepackage{mathtools}
\usepackage{subfigure}

\usepackage{bbm}
\usepackage[ruled,vlined]{algorithm2e}
\usepackage{color}

\newcommand{\etal}{\textit{et al}.}
\newcommand{\ie}{\textit{i}.\textit{e}.}
\newcommand{\eg}{\textit{e}.\textit{g}.}

\newcommand{\Tref}[1]{Table~\ref{#1}}
\newcommand{\Eref}[1]{Eq.~(\ref{#1})}
\newcommand{\Fref}[1]{Fig.~\ref{#1}}

\newcommand{\Cref}[1]{Chap.~\ref{#1}}

\usepackage{cite}

\begin{document}
%
\title{MCDAL: Maximum Classifier Discrepancy for Active Learning}
%
%
%

\author{Jae Won Cho$^*$, Student Member, IEEE,
        Dong-Jin Kim$^*$,  Member, IEEE,\\
        Yunjae Jung, Student Member, IEEE,
        In So Kweon, Member, IEEE
\thanks{$^*$ indicates equal contribution}
\thanks{Jae Won Cho, Yunjae Jung, and In So Kweon are with the Department
of Electrical Engineering, Korea Advanced Institute of Science and Technology, Daejeon,
Republic of Korea. (E-mail: \{chojw@kaist.ac.kr, yun9298a@gmail.com, iskweon77@kaist.ac.kr\})}
\thanks{Dong-Jin Kim is with the Electrical Engineering and Computer Science Department, University of California, Berkeley,
CA. (E-mail: djnjusa@gmail.com)}
}

%
%

\markboth{IEEE Transactions on Neural Networks and Learning Systems}%
{Shell \MakeLowercase{\textit{Cho et al.}}: IEEEtran.cls for IEEE Journals}
%



\maketitle

\begin{abstract}
Recent state-of-the-art active learning methods have mostly leveraged Generative Adversarial Networks (GAN) for sample acquisition; however, GAN is usually known to suffer from instability and sensitivity to hyper-parameters.
In contrast to these methods, 
we propose in this paper a novel active learning framework that we call Maximum Classifier Discrepancy for Active Learning (MCDAL) which takes the prediction discrepancies between multiple classifiers.
In particular, we utilize two auxiliary classification layers that learn tighter decision boundaries by maximizing the discrepancies among them.
Intuitively, the discrepancies in the auxiliary classification layers' predictions indicate the uncertainty in the prediction. 
In this regard, we propose a novel method to leverage the classifier discrepancies for the acquisition function for active learning.
We also provide an interpretation of our idea in relation to
existing GAN based active learning methods and domain adaptation frameworks.
Moreover, we empirically demonstrate the utility of our approach where the performance of our approach exceeds the state-of-the-art methods on several image classification and semantic segmentation datasets in active learning setups.\footnote{Our code is available at \url{https://github.com/chojw/Maximum-Classifier-Discrepancy-for-Active-Learning}}
\end{abstract}


%
\IEEEpeerreviewmaketitle

%
%
%
%

\section{Introduction}
\IEEEPARstart{I}{n}
recent years, deep learning has made great advancements with the help of large labeled datasets~\cite{deng2009imagenet} that have been handcrafted and labeled through various expensive methods~\cite{he2016deep,krizhevsky2012imagenet}. The cost of labeling and procuring such datasets have frequently pushed many researchers to find ways to minimize the need for methods to train powerful deep learning models that require fewer labels such as semi-supervised learning~\cite{tarvainen2017mean}, few-shot learning~\cite{snell2017prototypical}, or active learning~\cite{cohn1996active}.
Active learning explores a setting where there is a large dataset, but the budget to annotate the dataset is limited. 
In this setting, a small set of labeled samples is first used to train a model, then this trained model is used to decide which of the unlabeled samples would give the largest performance gain if given labels to. In recent years, active learning has been shown to be a practical approach to achieving high performance with relatively fewer samples in computer vision tasks such as image classification~\cite{beluch2018power,gal2017deep,sener2017active}, semantic segmentation~\cite{kuo2018cost,shin2021labor,yang2017suggestive}, and hyperspectral image classification~\cite{ahmad2020fuzziness,ahmad2020spatial}.

\begin{figure}[t]
\centering
   \includegraphics[width=1\linewidth]{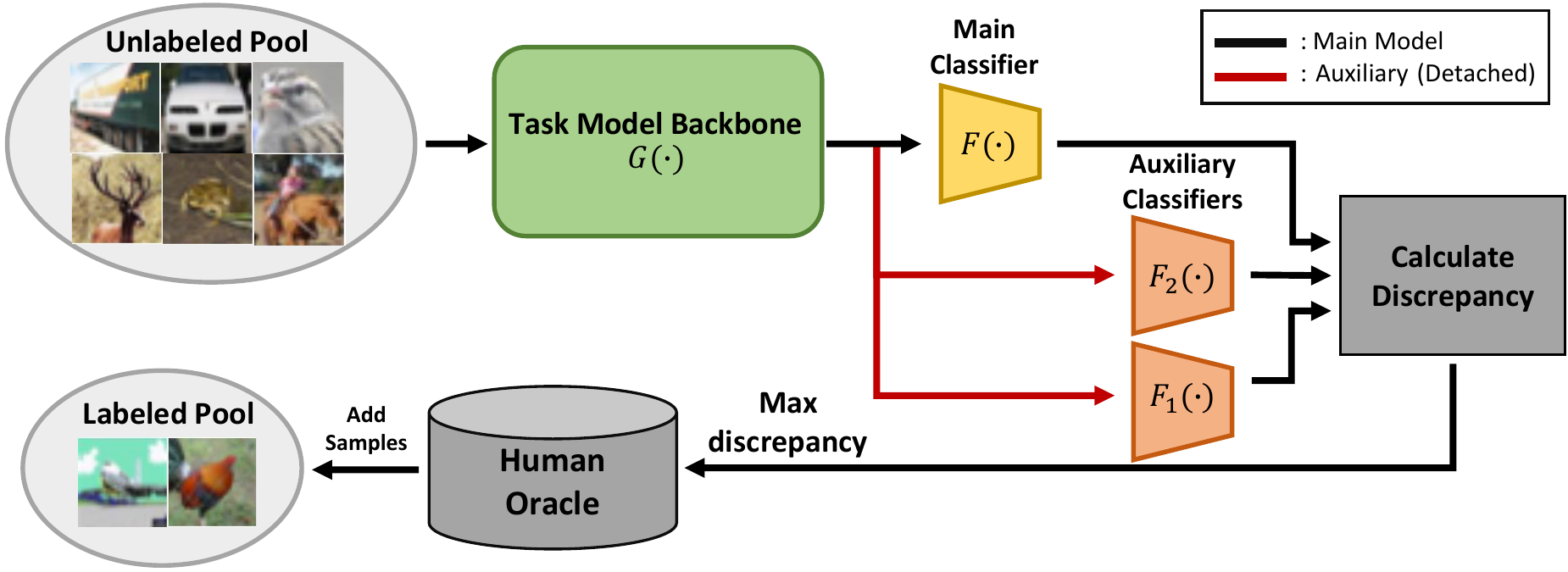}
   \caption{Overall architecture of our model. $G(\cdot)$ shows the ``Generator'' or the task model backbone, $F(\cdot)$ shows the ``main classifier'' (classification layer). $F\circ G(\cdot)$ represents off-the-shelf deep neural networks, and $F_1(\cdot)$ and $F_2(\cdot)$ are our auxiliary classification layers. In the proposed method MCDAL, we propose to collect unlabeled samples that shows the highest classifier discrepancy to add to a labeled pool.
   }
\label{fig:arch}
\end{figure}

Most recent works in active learning have started to use a Generative Adversarial Network~\cite{goodfellow2014generative} (GAN) based method where a Variational Auto-Encoder (VAE)~\cite{kingma2013auto} is used to train latent space representations of both the labeled and unlabeled datasets.
The VAE learns to fool a discriminator that both the labeled and unlabeled datasets are within the same pool. 
This discriminator is then used to measure how likely a sample is within the labeled set during the sampling stage, and samples that are not well represented in the labeled set are given labels to~\cite{sinha2019variational,shuo2020daal,zhang2020state}. Although these methods have shown promising performance, adding an auxiliary GAN based network to the existing task models is heavy and unstable~\cite{arjovsky2017wasserstein,kim2019image}. 
In addition, adversarial training is shown to be heavily sensitive to hyper-parameter choices between the discriminator and the generator~\cite{berthelot2017began}.

In order to tackle this issue, we propose a novel minimally intrusive and easy to implement method for training and sampling with less additional model parameters for active learning. 
In this paper, we introduce a novel ad hoc classifier \emph{discrepancy} based method for active learning that we call Maximum Classifier Discrepancy for Active Learning (MCDAL). 
To make our model as unobtrusive as possible, we leave the task model alone and add two new identical classification layers that are exactly the same as the task model's, making our model easy to implement. 
Note that these auxiliary classification layers do not effect the training of the task model (not a regularizer) as the classification branches are detached from the task model and are removed after the training process. In addition, as our method creates identical layers, our method can be readily applied to other tasks and datasets.
Then, using both the labeled and unlabeled dataset, we train the additional layers using both labeled cross-entropy loss and discrepancy loss in order for the ad hoc layers to mimic the ``main'' classification layer and to have a tighter decision boundary at the same time. 

After training our model in this manner, we leverage these auxiliary layers during the active learning sampling stage with our novel sampling method. 
Because the classification layers are trained to have tighter decision boundaries, intuitively, the samples that give the \emph{largest discrepancy} from the outputs of all the classifiers would be both uncertain and far from the labeled distribution.
Thus, we hypothesize that labeling these samples would give the greatest performance gain at each sampling stage. To the best of our knowledge, our work is the first work that leverages classifier discrepancy for sampling in active learning.

Our contribution in this work can be summarized as follows:
(1) We propose a novel active learning framework named Maximum Classifier Discrepancy for Active Learning (MCDAL) which is lightweight, fast, and easy to implement compared to the recent GAN based active learning methods.
(2) 
We provide an interpretation in several perspectives to understand the connection of MCDAL with GAN based active learning methods and domain adaptation frameworks.
(3) Extensive experiments and analysis consistently show the performance surpassing recent state-of-the-art active learning works with a noticeable margin.

\label{sec.intro}

\section{Related Work}


\noindent\textbf{Active leaning for deep vision.}
Active learning has been widely studied with several different approaches being introduced in recent years to tackle the data issue. Several methods have been proposed to tackle the data issue such as Bayesian Dropout~\cite{xie2021advanced} and interactive learning~\cite{fails2003interactivelearning,ahmad2019interactive,ho2020deep}, however in this work we focus on the task of active learning.
Classical active learning approaches either use pool-based or query synthesis methods. In query synthesizing approaches, generative models are used to find the most informative samples~\cite{mahapatra2018efficient,mayer2020adversarial,zhu2017generative}. 
In pool-based methods, there are also several categories: uncertainty-based ~\cite{beluch2018power,collins2008towards,joshi2009multi,kim2021single,wang2016cost,yoo2019learning}, representation-based~\cite{sener2017active}, and more recently a combination of these  two~\cite{sinha2019variational,zhang2020state} and our work belongs here.

In recent years, pool-based active learning has been successfully applied in many deep vision tasks such as image recognition~\cite{sener2017active,yoo2019learning}, 
object detection~\cite{aghdam2019active}, and semantic segmentation~\cite{agarwal2020contextual}.
Meanwhile, theoretical dropout-based frameworks have also been used to measure uncertainty~\cite{gal2017deep,kendall2017uncertainties}.
More recently, works have started to use latent space representations with the aid of Variational Auto-Encoders (VAE)~\cite{kingma2013auto} in addition to adversarial training methods~\cite{goodfellow2014generative} to find the informativeness of unlabeled samples~\cite{sinha2019variational,shuo2020daal,zhang2020state}. Although these methods have shown to be promising, we choose to turn our attention away from Generative Adversarial Network (GAN)~\cite{goodfellow2014generative} based methods due to the instability and complexity in training and increase in network parameters.
Instead, we focus on classifier discrepancy based training methods which introduce little addition parameters to the network. In the world of using additional classifiers, works such as such as~\cite{chang2021your} incorporate additional classifiers for fine-grained classification.

Unlike ensemble based method~\cite{beluch2018power,gal2016dropout}, where~\cite{beluch2018power} uses several randomly initialized models for ensembling while~\cite{gal2016dropout} uses different dropout masks from a single \emph{model}, creating additional computation time as several forward passes are required, we share a backbone model and add auxiliary fully connected \emph{layers}, while introducing significantly less parameters, being extremely lightweight, and surpassing the state-of-the-art performance.

\noindent\textbf{Maximum classifier discrepancy in domain adaptation.}
The goal of domain adaptation is to improve the testing performance on an unlabeled target domain while a model is trained on a related yet different source domain. 
Recent GAN~\cite{goodfellow2014generative} based methods~\cite{hoffman2018cycada,long2018conditional,tzeng2017adversarial,saito2018adversarial} have leveraged a discriminator to distinguish source and target domain by trying to reduce the discrepancy of the distributions based on the theory proposed by~\cite{ben2007analysis}.
However, distribution aligning methods using GAN do
not consider the relationship between target samples and \emph{decision boundaries}. 
To tackle these problems, Saito~\etal~\cite{saito2018maximum} proposed an approach to directly use task-specific classifiers as a discriminator.
In order to achieve this, 
the authors implement a two-classifier system with discrepancy loss to maximize the disagreement between the classifiers, leading to tighter decision boundaries.
Also, they train a generator to minimize this disagreement to align the source and target features given the tight decision boundaries. 
The idea of maximizing the classifier discrepancy is utilized in other tasks that require distribution alignment such as out-of-distribution detection~\cite{yu2019unsupervised}.

Similarly, we tackle the problem of existing GAN based active learning approaches and propose a novel classifier discrepancy based active learning method which only requires task-specific classification layers.
Motivated by the goal of domain adaptation, we propose to select unlabeled samples that show the largest discrepancy when compared to the distribution of the labeled dataset in the sample acquisition stage.
Unlike~\cite{saito2018maximum}, we introduce two auxiliary classification layers alongside the main classifier without training the generator. Instead, in the sample acquisition stage, we select the unlabeled samples that have maximum classifier discrepancy.
Although \cite{fu2021agreement} tried to leverage classifier discrepancy maximization as a semi-supervised learning technique instead of active learning, we propose a novel active learning method to directly leverage classifier discrepancy.


\label{sec.related}

\section{Proposed Method}

In this section, we introduce our method, Maximum Classifier Discrepancy for Active Learning (MCDAL) in detail and the methods for both training and sampling. For clarity, we first show~\Tref{tab:Notation} with the list of notations used in this work.

\begin{table}[ht]
\begin{center}
    \caption{Notations and their descriptions used throughout this work.}
    \resizebox{1\linewidth}{!}{
    \begin{tabular}{l l}
        \hline
        Notation & Description \\
        \hline
        $\mathcal{D}\textsubscript{L} = \{x\}$ & Labeled set\\
        $\mathcal{D}\textsubscript{U} = \{(x,y)\}$ & Unlabeled set \\
        $C$ & Total number of classes\\
        $F(\cdot)$ & Backbone architecture \\
        $G(\cdot)$ & Main classification layer\\
        $\theta$ & Parameter for $F \circ G(\cdot)$\\
        $F_1(\cdot)$ \& $F_2(\cdot)$ & Auxiliary classification layers with parameter $\theta_1$ \& $\theta_2$\\
        $p(y|x)$& Output probability prediction from the main classifiers\\
        $p_1(y|x)$ \& $p_2(y|x)$ & Probability prediction from the auxiliary classifiers\\
        $S(\cdot)$ & Acquisition function\\
        $D(x)$ & Total discrepancy metric given an image\\
        $d(\cdot,\cdot)$ & L1 distance \\
        $\eta$ & Learning rate \\
        \hline
        \end{tabular}
    }
    \label{tab:Notation}
\end{center}
\end{table}

\subsection{Problem Definition : Active Learning}
Given an image $x$, the general aim of active learning is finding the best sample acquisition function $S(x)$ that assigns the highest scores to the most \emph{informative} samples, which can be seen as samples that increase the performance gain the most when labeled, given a fixed backbone architecture. 
In this setup, a labeled set $\mathcal{D}\textsubscript{L}$ is first used to train a model. 
After the model converges, the model is used to sample from the unlabeled set $\mathcal{D}\textsubscript{U}$. 
Finally, the chosen samples are moved from the unlabeled set $\mathcal{D}\textsubscript{U}$ to the labeled set $\mathcal{D}\textsubscript{L}$ and are given labels.
This step (otherwise called a \emph{stage}) is repeated several times depending on the experimental setup. 
Recently, Generative Adversarial Networks (GAN) based methods~\cite{sinha2019variational,zhang2020state} have shown state-of-the-art active learning performances by finding the unlabeled samples that has the largest latent space gap with the labeled sample distribution.
However, GAN based methods usually suffer from difficulty in training, such as instability or sensitivity to hyper-parameters between the discriminator and the generator~\cite{arjovsky2017wasserstein,berthelot2017began,kim2019image}.
In light of this, we define a novel acquisition function that takes the \emph{discrepancy} of the classifiers into account.
Our approach utilizes multiple auxiliary classification layers and measures the discrepancy between the outputs of the classification layers.

\subsection{Network Architecture}

Let us denote $G(\cdot)$ as the network backbone that generates features from an input $x$.
Then, with the ``main'' classification layer $F(\cdot)$, a class prediction is computed ($p(y|x)=F\circ G(x)\in \mathbb{R}^C$ where $C$ is the total number of classes).
The combined network $F\circ G(\cdot)$ can be implemented with typical image classifiers~\cite{he2016deep,krizhevsky2012imagenet}.
The main model is trained with cross-entropy loss with the ground truth label $y$:
\begin{equation}
    \begin{split}
    &\min_{F,G} \mathcal{L}_{CE}(F,G), \text{ where }\\ 
    &\mathcal{L}_{CE} = \mathop{\mathbb{E}}_{(x,y)\in \mathcal{D}\textsubscript{L}}\bigg[-\sum_{c=1}^C \mathbbm{1}[c=y]\log p^c(y|x)\bigg],
    \end{split}
\end{equation}
where $p^c(y|x)$ denotes a probability element of $p$ for class $c$
and $\mathbbm{1}[a]$ is the indicator function, which is 1 if predicate $a$ is true and 0 otherwise.

In order to measure the classifier discrepancy, we introduce two additional classification layers $F_1(\cdot)$ and $F_2(\cdot)$. 
The extra layers introduced share the identical shape as $F(\cdot)$, meaning that our method can be easily applied to any network and any task. In addition, note that we detach the output from $G(\cdot)$ before we pass it through $F_1(\cdot)$ and $F_2(\cdot)$ in order to not affect the task-model and to objectively test our method. Although these classification layers are still trained with the typical cross-entropy loss $\mathcal{L}_{CE}$, we do not consider the performance of the auxiliary classifiers as their main goal is not to classify for the task correctly, but to learn tighter decision boundaries for sample acquisition.

\begin{algorithm}[ht]
\SetAlgoLined
\KwIn{Labeled pool $\mathcal{D}\textsubscript{L}$, and unlabeled pool $\mathcal{D}\textsubscript{U}$, Initialized models where $F\circ G(\cdot), F_1(\cdot), \text{ and }F_2(\cdot)$ with parameters $\theta$,$\theta_1$,\text{ and }$\theta_2$ respectively}
\KwIn{Hyper-parameters : learning rate $\eta$, maximum number of epochs $MaxEpoch$}
\KwOut{$F\circ G(\cdot), F_1(\cdot), \text{ and }F_2(\cdot)$ with trained $\theta$,$\theta_1$,\text{ and }$\theta_2$}
\Begin{
    \For{$e=1$ to MaxEpoch}{
    sample $(x_l,y_l)\in\mathcal{D}\textsubscript{L}$\;
    Compute $\mathcal{L}_{CE}$\;
    Update the model parameters:\\
    $\theta \leftarrow \theta - \eta \nabla \mathcal{L}_{CE}$\\
    $\theta_1 \leftarrow \theta_1 - \eta \nabla \mathcal{L}_{CE}$\\
    $\theta_2 \leftarrow \theta_2 - \eta \nabla \mathcal{L}_{CE}$\\
    sample $(x_u)\in\mathcal{D}\textsubscript{U}$\;
    Compute $\mathcal{L}_{dis}$\;
    Update the model parameters:\\
    $\theta_1 \leftarrow \theta_1 + \eta \nabla \mathcal{L}_{dis}$\\
    $\theta_2 \leftarrow \theta_2 + \eta \nabla \mathcal{L}_{dis}$\\
    }
}
\caption{Training MCDAL}
\label{alg:training}
\end{algorithm}

\subsection{Training with Discrepancy Losses}

Let us denote $p_1$ and $p_2$ as the output probability computed from the auxiliary classification layers $F_1(\cdot)$ and $F_2(\cdot)$, $p_1 (y|x)=F_1\circ G(x)\in \mathbb{R}^C$, $p_2 (y|x) =F_2\circ G(x)\in \mathbb{R}^C$.
The auxiliary layers $F_1(\cdot)$ and $F_2(\cdot)$ are trained with the following objective:
\begin{equation}
    \min_{F_1,F_2} \mathcal{L}_{CE}(F_1,G) + \mathcal{L}_{CE}(F_2,G) - \mathcal{L}_{dis}(F_1,F_2,G).
    \label{eqn:unlabeled}
\end{equation}
The objective function consists of the cross-entropy loss and the additional discrepancy loss which is defined as follows:
\begin{equation}
    \mathcal{L}_{dis} = \mathop{\mathbb{E}}_{x\in \mathcal{D}\textsubscript{U}}\bigg[d(p_1,p) + d(p_2,p) + d(p_1,p_2)\bigg],
    \label{eqn:dis_loss}
\end{equation}
which is the sum of the discrepancy values between all pairs of probabilities.

As a discrepancy measure, we utilize the absolute values of the difference between the classifiers’ output probabilities (L1 distance between the probability vectors):
\begin{equation}
    d(p_1,p_2) = \frac{1}{C}\sum_{c=1}^C{\Big| p_1^c(y|x) - p_2^c(y|x) \Big|},
    \label{eqn:l1dist}
\end{equation}
where the $p_1^c(y|x)$ and $p_2^c(y|x)$ denote probability elements of $p_1$ and $p_2$ for class $c$ respectively. 
The choice for L1-distance is based on the theory proposed by~\cite{ben2010theory}.
We tried other distance or divergence metrics such as L2 and KL-Divergence (KL) in the experiment in \Fref{fig:distance}, and we empirically find that L1 is the best option in our given setting.
Unlike other works that train a \emph{separate} feature generator~\cite{saito2018maximum,yu2019unsupervised}, note that we do not train our feature generator $G(\cdot)$ with the discrepancy loss as the scope of active learning is sampling and not training the task model with additional methods. Since the discrepancy loss is computed with unlabeled samples, including the discrepancy loss into the feature generator would result in an unfair comparison to other active learning methods as this can be viewed as semi-supervised training. We show our pseudo-code for training in Algorithm 1 for ease of understanding.

\begin{figure}[t]
\centering
   \includegraphics[width=1\linewidth]{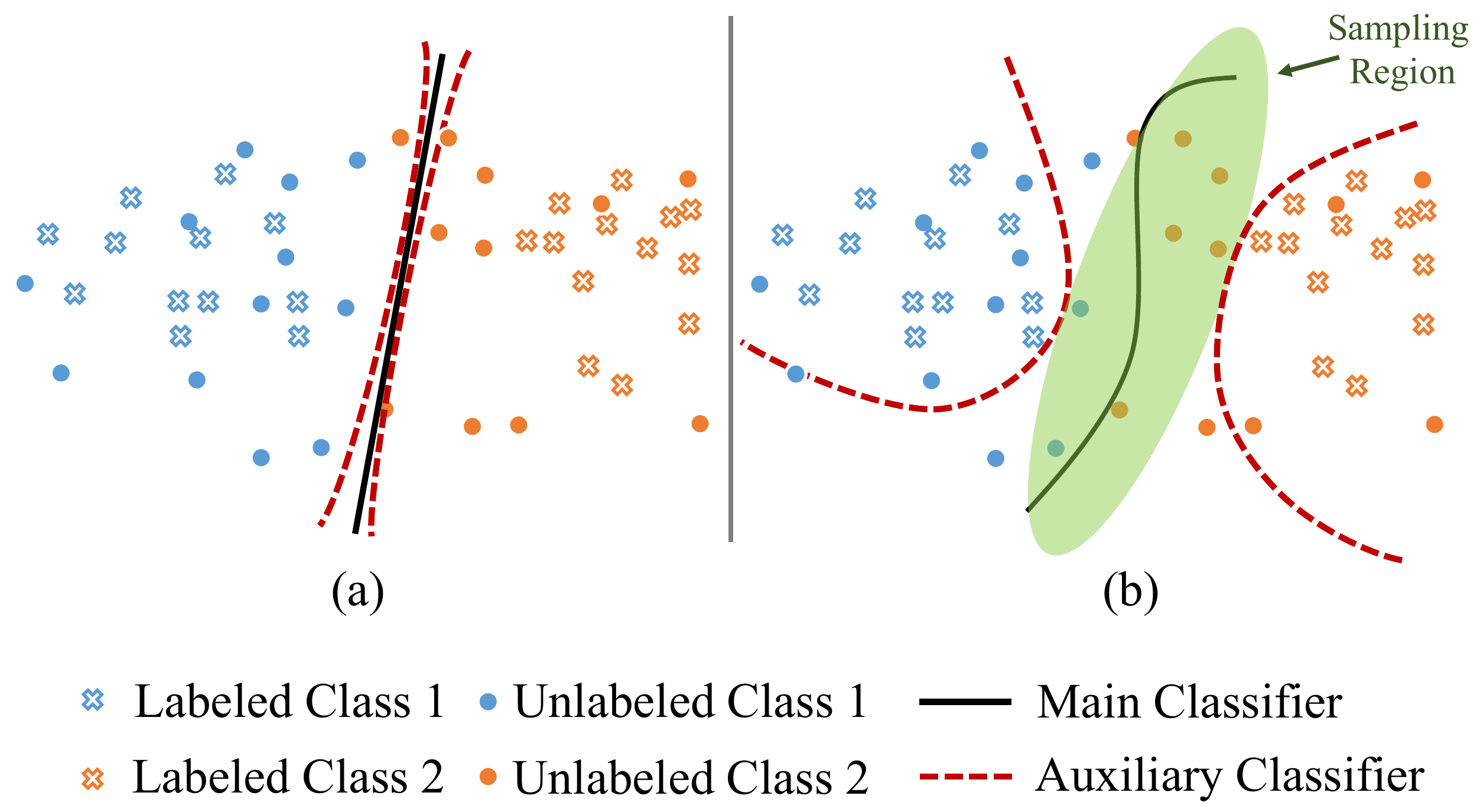}
   \caption{Illustration of training our auxiliary classification layers based on the discrepancy in feature space. 
   The two dotted red lines each are for each of the auxiliary classifiers.
   (a) shows the early stages of the training where all the classifiers are similar and (b) shows when after training with the discrepancy loss (\Eref{eqn:dis_loss}), the classifiers are pushed apart to maximize discrepancy and forms a space that we to consider the ``sampling region.''}
\label{fig:illustration}
\end{figure}

By training the auxiliary layers to maximize the discrepancy between their respective outputs, the decision boundaries of the classifiers become tighter as illustrated in \Fref{fig:illustration}, resulting in the region between the auxiliary decision boundaries in the feature space to become larger.
As the supervised cross-entropy loss is also applied to train the auxiliary layers, we expect most of the labeled samples to be located within the boundaries, whereas some of the labeled samples would be located out of the boundaries (we call this region the ``sampling region''). 
In the sample acquisition stage, our key idea is to collect the unlabeled samples that are located 
in this sampling region.
The detailed acquisition process is depicted in the following subsection.

\subsection{Sampling with MCDAL}
\label{sec.sampling}

By maximizing the discrepancy between auxiliary classification layers, we obtain two additional tight decision boundaries which means that we have a large region between the two decision boundaries (we call this region the ``sampling region'').
The unlabeled samples that are (1) difficult to train and that are (2) far from the labeled data distribution will be located 
in this ``sampling region''
which meets the condition that would be helpful if labeled.
In this regard, we again utilize the discrepancy between the auxiliary classification layers by defining the sample acquisition function $S(\cdot)$ as $S(x_u) = D(x_u)$ 
for $x_u\in\mathcal{D}\textsubscript{U}$, where $D(x) = d(p_1,p_2)$ (note $p_1$ and $p_2$ in this equation are interchangeable with $p$ where needed).
In order to further leverage the difference between the labeled set and unlabeled set, 
our final acquisition function compares the average classifier discrepancy of the labeled samples:
\begin{equation}
    S(x_u) = \Bigg| D(x_u) - \frac{1}{|\mathcal{D}\textsubscript{L}|}\sum_{x_l\in\mathcal{D}\textsubscript{L}}{D(x_l)} \Bigg|, 
    \text{for } x_u\in\mathcal{D}\textsubscript{U}.
    \label{eqn:final}
\end{equation}

This in turn measures how large the classifier discrepancies for the given unlabeled samples are compared to the average classifier discrepancies of the labeled samples. 
For the final design of $S(\cdot)$, we use $D(x) = d(p_1,p_2) + d(p,p_1) + d(p,p_2)$ to leverage the discrepancy between all present classifiers and include the pseudo-code in Algorithm 2 for ease of understanding.
Although we consider using the discrepancy value directly  (\ie,~$S(\cdot) = D(\cdot)$), we do not do so and explain in further detail in the following subsection. 

\begin{algorithm}[ht]
\SetAlgoLined
\KwIn{Labeled pool $\mathcal{D}\textsubscript{L}$, Unlabeled pool $\mathcal{D}\textsubscript{U}$, Number of samples collecting for each stage $b$, Models $F\circ G(\cdot), F_1(\cdot), \text{ and }F_2(\cdot)$  }
\KwOut{Updated $\mathcal{D}\textsubscript{L}$ and $\mathcal{D}\textsubscript{U}$}
\Begin{
    From $\mathcal{D}\textsubscript{U}$, collect a set of samples $\{x_s\}$ with 
    $\max_b S(x_u) = \max_b{  \Bigg| D(x_u) - \frac{1}{|\mathcal{D}_L|}\sum_{x_l\in\mathcal{D}_L}{D(x_l)} \Bigg|},$\ 
    where $ D(x) = d(p_1,p_2) + d(p,p_1) + d(p,p_2),$\
    where $ d(p_1,p_2) = \frac{1}{C}\sum_{c=1}^C{|p_1^c(y|x) - p_2^c(y|x)|}$;
    $\{(x_s,y_s)\} = ORACLE(\{x_s\})$\;
    $\mathcal{D}\textsubscript{L} \leftarrow \mathcal{D}\textsubscript{L} \cup \{(x_s,y_s)\}$\;
    $\mathcal{D}\textsubscript{U} \leftarrow \mathcal{D}\textsubscript{U} - \{x_s\}$\;
}
 \caption{Sampling with MCDAL}
 \label{alg:sampling}
\end{algorithm}

\subsection{Discussion}

\noindent\textbf{Connection with GAN based approaches.}
The goal of GAN based active learning works~\cite{sinha2019variational,zhang2020state} is to train a discriminator to distinguish labeled and unlabeled samples so that they can collect the unlabeled samples that are the furthest from (not well represented in) the distribution of the labeled data.
However, these approaches do not explicitly consider the class-wise behavior of the samples.
Sinha~\etal~\cite{sinha2019variational} does not consider the model's final output, and Zhang~\etal~\cite{zhang2020state} only adds
output uncertainty into the GAN model, but they also do not leverage class-aware behavior based in their uncertainty measure.
In our case, what we sample in the acquisition stage are the unlabeled samples that are both uncertain (as the uncertainty is large in the ``sampling region'') and far from the distribution of the labeled data (as the labeled samples are trained to be outside the ``sampling region''). Therefore, our approach has a more general effect in comparison to 
the existing GAN based active learning approaches.

\noindent\textbf{Domain adaptation perspective.}
As classifier discrepancy has been used in several domain adaptation research~\cite{saito2018maximum}, our approach can also be interpreted in the domain adaptation perspective.
In particular, the labeled and unlabeled data pools in active learning can be seen as source and target domain in domain adaptation, respectively.

Since the classifier discrepancy based domain adaptation works~\cite{saito2018maximum} are motivated by the theory in domain adaptation proposed by Ben-David~\etal~\cite{ben2010theory}, we show the relationship between our method and the theory in this section.
Ben-David~\etal~\cite{ben2010theory} proposed the theory that bounds the expected error on the target samples, $R_\mathcal{T}(h)$, by using three terms: (1) expected error on the source domain, $R_\mathcal{S}(h)$; (2)
$\mathcal{H}\Delta \mathcal{H}$-distance ($d_{\mathcal{H}\Delta \mathcal{H}}(\mathcal{S},\mathcal{T})$), which is measured as the discrepancy between two classifiers; and (3) the shared error of the ideal joint hypothesis, $\lambda$. 
$\mathcal{S}$ and $\mathcal{T}$ denote source and target datasets respectively. 
The theory can be explained as follows:
let $\mathcal{H}$ be the hypothesis class. Given two domains S and T, we have
\begin{equation}
\centering
    \begin{split}
    \forall h \in \mathcal{H}, R_\mathcal{T}(h) &\leq R_\mathcal{S}(h) + \frac{1}{2}d_{\mathcal{H}\Delta \mathcal{H}}(\mathcal{S},\mathcal{T}) + \lambda,\\
    \end{split}
\end{equation}
where $d_{\mathcal{H}\Delta \mathcal{H}}(\mathcal{S},\mathcal{T})$ is defined as:
\begin{equation}
2 \sup_{(h,h')\in \mathcal{H}^2} \Bigg| \mathop{\mathbb{E}}_{x\in \mathcal{S}} \mathbbm{1}[h(x) \neq  h'(x)] - \mathop{\mathbb{E}}_{x\in \mathcal{T}} \mathbbm{1}[h(x) \neq  h'(x)] \Bigg|.
\end{equation}
Here, $\lambda = \min|R_\mathcal{S}(h) + R_\mathcal{T}(h)|$,
$R_\mathcal{T}(h)$ is the error of hypothesis $h$ on the target domain, and $R_\mathcal{S}(h)$ is the corresponding error on the source domain. 
$\lambda$ is a constant 
which is considered sufficiently
low to achieve an accurate adaptation. 
We will show the relationship between our method and $\mathcal{H}\Delta \mathcal{H}$-distance.

In our case $h$ and $h'$ share the feature extractor $G(\cdot)$,
we decompose the hypothesis $h$ into $G(\cdot)$ and $F_1(\cdot)$, and $h'$ into $G(\cdot)$ and $F_2(\cdot)$. $G(\cdot)$, $F_1(\cdot)$ and $F_2(\cdot)$ correspond to the network in our method. 
Substituting
these notations into the $\sup_{(h,h')\in \mathcal{H}^2}|\mathbb{E}_{x\in \mathcal{S}} \mathbbm{1}[h(x) \neq  h'(x)] - \mathbb{E}_{x\in \mathcal{T}} \mathbbm{1}[h(x) \neq  h'(x)]|$ and for fixed $G(\cdot)$, the term 
becomes:
\begin{equation}
    \begin{split}
    \sup_{F_1,F_2}& \Bigg| \mathop{\mathbb{E}}_{x\in \mathcal{S}} \mathbbm{1}[F_1 \circ G(x) \neq  F_2 \circ G(x)] \\
    &- \mathop{\mathbb{E}}_{x\in \mathcal{T}} \mathbbm{1}[F_1 \circ G(x) \neq  F_2 \circ G(x)] \Bigg|.
    \end{split}
\end{equation}
Furthermore, if we replace the expectations $\mathbb{E}_{X\in \mathcal{S}}[\cdot]$ and $\mathbb{E}_{X\in \mathcal{T}}[\cdot]$ as the empirical means in the labeled and unlabeled datasets $\mathcal{D}\textsubscript{L}$ and $\mathcal{D}\textsubscript{U}$ respectively, and if we replace $\sup$ with $\max$, we obtain:
\begin{equation}
   \begin{split}
    \max_{F_1,F_2} & \Bigg|
    \frac{1}{|\mathcal{D}\textsubscript{L}|}\sum_{x_l\in\mathcal{D}\textsubscript{L}}{\mathbbm{1}[F_1 \circ G(x_l) \neq  F_2 \circ G(x_l)]} \\
    &-  \frac{1}{|\mathcal{D}\textsubscript{U}|}\sum_{x_u\in\mathcal{D}\textsubscript{U}}{\mathbbm{1}[F_1 \circ G(x_u) \neq  F_2 \circ G(x_u)]}   \Bigg|,
    \end{split}
    \label{eqn:difference}
\end{equation}
which is similar to
our final acquisition function in
\Eref{eqn:final}.
Moreover, in order to increase the tractability in training stage, we introduce further assumptions.
As $h$ and $h'$ are expected to classify labeled samples perfectly, we can assume the term $\frac{1}{|\mathcal{D}\textsubscript{L}|}\sum_{x_l\in\mathcal{D}\textsubscript{L}}{\mathbbm{1}[F_1 \circ G(x_l) \neq  F_2 \circ G(x_l)]}$ to be close to zero. 
In other words, $h$ and $h'$ should agree on their predictions on labeled samples.
Thus, \Eref{eqn:difference} can be approximated as: 
\begin{equation}
    \centering
    \max_{F_1,F_2}\frac{1}{|\mathcal{D}\textsubscript{U}|}\sum_{x_u\in\mathcal{D}\textsubscript{U}}{\mathbbm{1}[F_1 \circ G(x_u) \neq  F_2 \circ G(x_u)]}.
    \label{eqn:approximated}
\end{equation}
This equation is similar to the optimization problem we solve in our discrepancy loss
in \Eref{eqn:dis_loss}, where classification layers are trained to maximize their discrepancy on unlabeled samples. 
Although we must train all branches to minimize the classification loss on labeled training data, we can see the connection to the theory proposed by~\cite{ben2010theory}.

While \cite{saito2018maximum} additionally requires minimizing the equation similar to \Eref{eqn:approximated} with respect to $G(\cdot)$, which makes the optimization unstable, we do not minimize the approximated objective \Eref{eqn:approximated} because it is not fair to optimize $G(\cdot)$ with the loss from unlabeled samples as previously mentioned.
Instead, through our acquisition function~\Eref{eqn:final}, we indirectly minimize the non-approximated objective \Eref{eqn:difference} by sampling the unlabeled sample that gives a large difference of discrepancy with the labeled data in the sample acquisition stage, and learn from the sample with supervised cross-entropy loss (which corresponds to the formulation in our final acquisition function in \Eref{eqn:final}).

\noindent\textbf{Semi-supervised learning perspective.} 
Our approach has similar motivation to that of consistency based semi-supervised learning works~\cite{qiao2018deep,tarvainen2017mean} which is to utilize the difference between the multiple outputs when training with unlabeled samples.
The goal of consistency based semi-supervised learning is to directly reducing the discrepancy between the model outputs with different conditions (\eg,~augmentations) for the unlabeled samples:
\begin{equation}
    \min_{F,G} \mathbb{E}_{x\in \mathcal{D}\textsubscript{U}}[d(p_1,p_2)].
\end{equation}
On the other hand, as our task is active learning, we do not apply additional losses other than supervised cross-entropy loss for fairness of testing.
Instead, we find the unlabeled training samples that have large discrepancy between the outputs from different branches so that we can exploit that samples with a supervised loss.
Therefore, we verify that our approach theoretically in the perspective of well-explored semi-supervised learning approaches.

\label{sec.method}

\section{Experiments}

In this section, we explain the implementation details and the experiments performed while providing detailed discussions of our results. We evaluate MCDAL against various recent state-of-the-art methods in addition to a random annotation oracle on the image classification and segmentation task. For our active learning setup, we start with an initial pool of 10\% of the training set. At the sampling stage, the oracle annotates from the unlabeled set, and each sampling stage increases the dataset by 5\%, and we repeat this step until the final stage where our labeled data consists of a total of 40\% of the training set.
To verify the performance of our sampling algorithm, we use the average of five runs and initialize the task model at every stage.

\subsection{Image Classification}

\noindent{\textbf{Datasets.}} For the image classification task, we test our method on the classical image classification datasets CIFAR-10~\cite{cifar10}, CIFAR-100~\cite{cifar10}, and Caltech-101~\cite{fei2006one} following the footsteps of some recent works. 
CIFAR-10 and CIFAR-100 consists of 50,000 training images and 10,000 test images. CIFAR-10 has 6,000 images per class while CIFAR-100 has 600 images per class. Caltech-101 has a total of 9,146 images with 101 classes and 1 background class. Each class in Caltech-101 has a different number of images ranging from 40 to 800 and we believe that this setup shows a more real-world-like setup. These different distributions of datasets show the different situations of active learning and how the different algorithms perform under these varying dataset conditions.

\begin{figure}[t]
\centering
  \includegraphics[width=.95\linewidth]{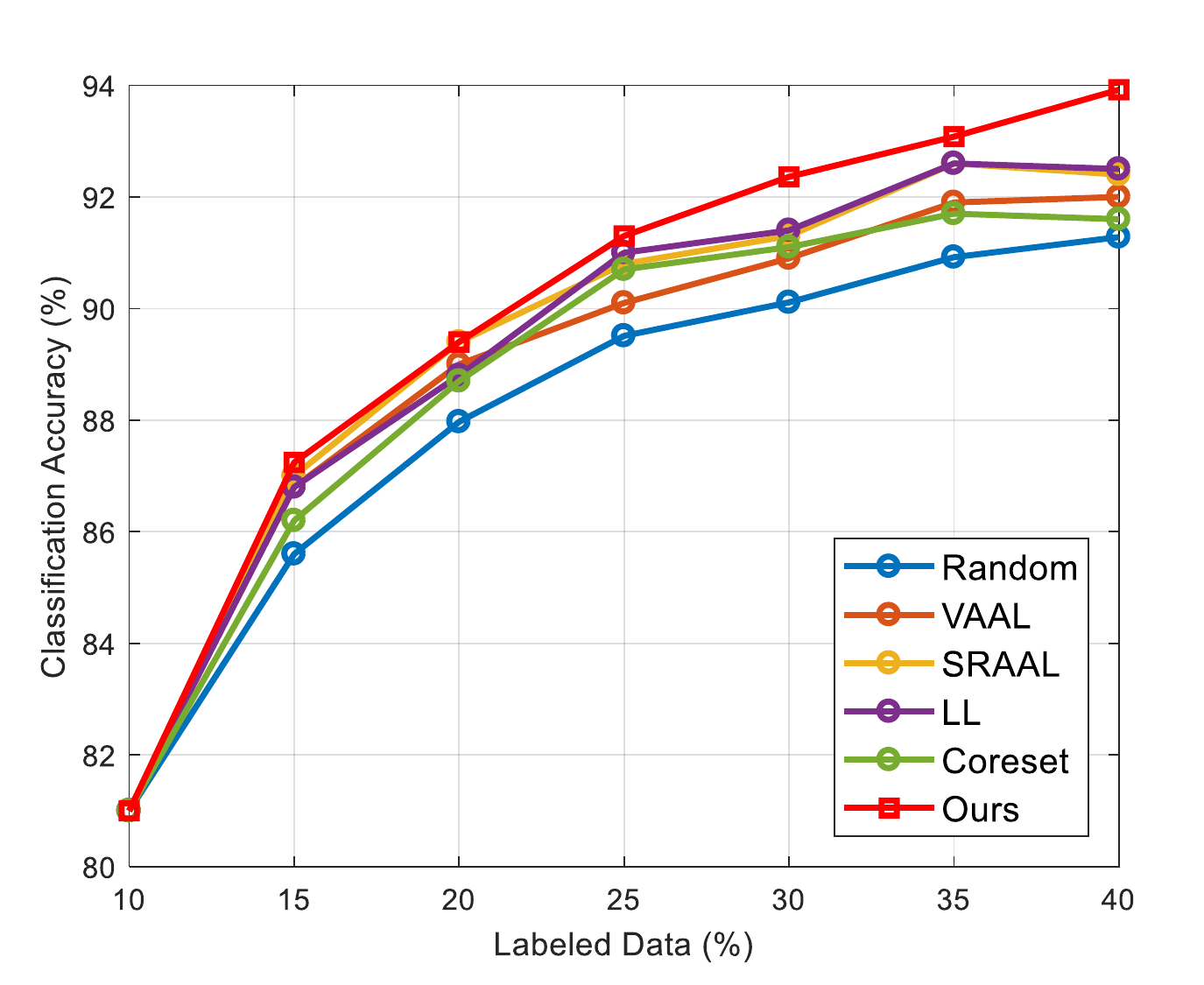}
  \caption{Comparison with existing baselines on CIFAR-10. Our method, MCDAL shows the best performance compared to the state-of-the-art methods in a noticeable margin, especially in the late stages.}
\label{fig:cifar10}
\end{figure}

\noindent{\textbf{Comparison baselines.}} We compare MCDAL to various recent-state-of-the-art approaches including Core-set~\cite{sener2017active}, Learning Loss (LL)~\cite{yoo2019learning}, VAAL~\cite{sinha2019variational}, and SRAAL~\cite{zhang2020state}. Although DAAL~\cite{shuo2020daal} is the most recent work, the implementation details in the paper for their VAE and discriminator is not clear enough to implement with no official repository, in addition to the performance being similar to SRAAL. 
Therefore, we do not compare to DAAL in this paper. We also include random sampling as a baseline. As most of the baselines do not have an official repository, we implement the baselines to the best of our ability using the information present in the respective papers.
We do not compare with classical methods (MC-dropout~\cite{gal2016dropout}, DBAL~\cite{gal2017deep}, Ensemble~\cite{beluch2018power}) in this paper as several previous works~\cite{sinha2019variational,zhang2020state} have shown that the classical method shown similar or worse performance in comparison to the Random sampling approach.

\noindent{\textbf{Implementation details.}} 
Following recent active learning works such as~\cite{yoo2019learning,zhang2020state}, we also adopt the Resnet-18~\cite{he2016deep} architecture and use a publicly available code~\footnote{https://github.com/weiaicunzai/pytorch-cifar100} as the base architecture for all image classification tasks and use the same image augmentation strategy from the aforementioned code. Since we use Resnet-18, we use the same classification layer (final linear layer)
for the respective tasks.
The features that pass through the auxiliary classification branches are detached so as not to affect the task model. We use Stochastic Gradient Descent as an optimizer for both the task model and auxiliary layers with the same learning rate of $1\times10^{-1}$ with a multi-step learning rate scheduler with a decay of $2\times10^{-1}$ at 30\%, 60\%, 80\% of the total training epochs. We train the CIFAR-10 and CIFAR-100 for 100 epochs and train Caltech-101 for 50 epochs as it is a much smaller dataset. We use budget sizes 2500 for CIFAR-10 and CIFAR-100, and 450 for Caltech-101.

\begin{figure}[t]
\centering
  \includegraphics[width=.95\linewidth]{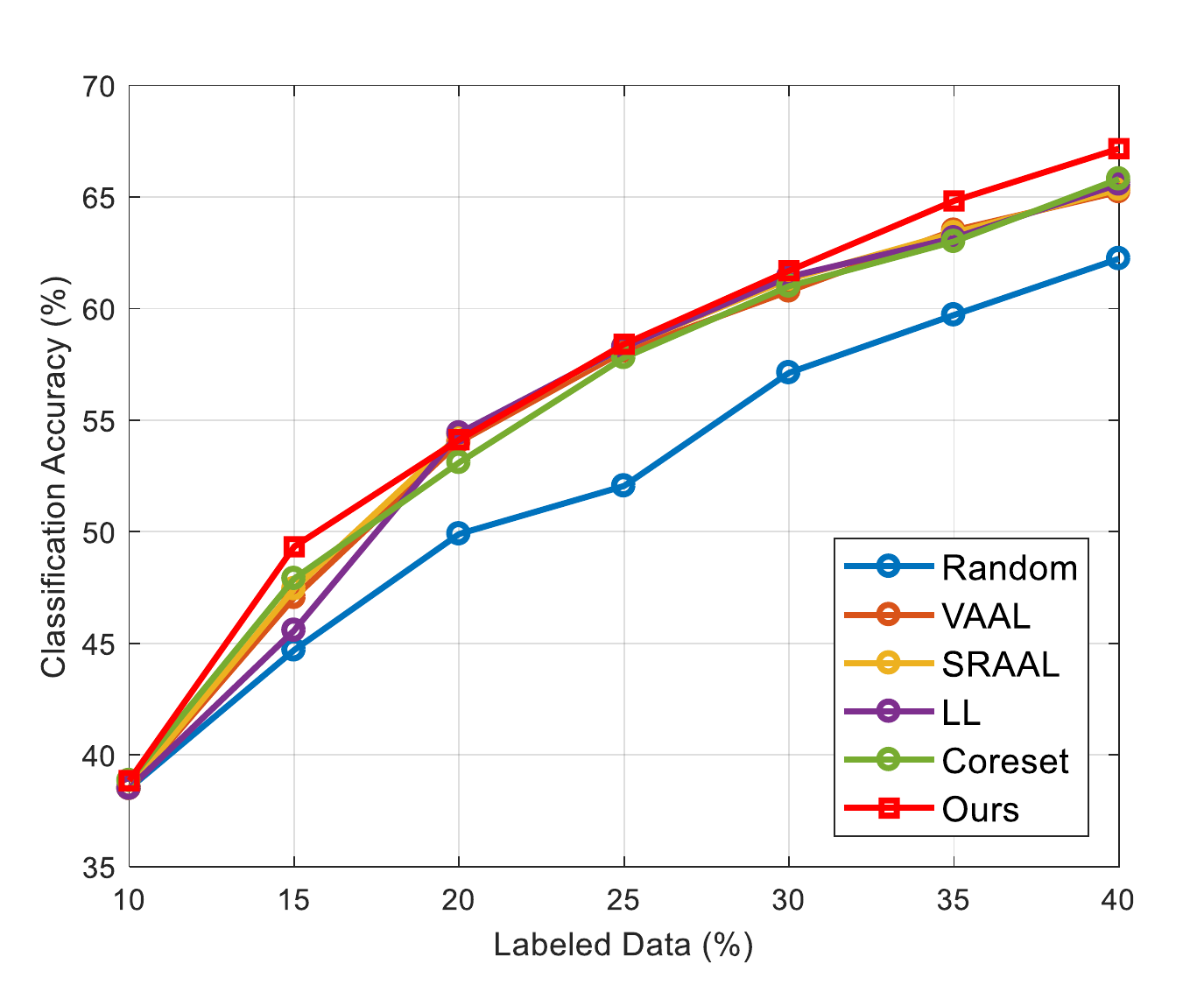}
  \caption{Comparison with existing baselines on CIFAR-100. Our method, MCDAL shows the best performance compared to the state-of-the-art methods overall.}
\label{fig:cifar100}
\end{figure}


\noindent\textbf{Performance on CIFAR-10.}
\Fref{fig:cifar10} shows the performance of MCDAL in comparison to recent state-of-the-art methods.
Among the baseline methods, SRAAL shows the highest performance in the early stage while LL shows the best performance in the late stage.
However, our method, MCDAL shows the best performance compared to the state-of-the-art methods in all the stages with a noticeable margin, especially in the late stages.
In our setting, using 100\% of the labeled dataset gives 94.51\% accuracy and MCDAL shows the highest performance of 93.92\% with standard deviation of $\pm0.24\%$ at 40\% labeled, which is only 0.59\% off. 
Although in the early stages of active learning VAAL, SRAAL, and LL shows similar performance as MCDAL, MCDAL consistently outperforms at all stages and shows large performance gain at the final stages. 
This is worth noting because the performances of other methods start to saturate in the later stages whereas MCDAL shows a continuously increasing trend.

\noindent\textbf{Performance on CIFAR-100.}
As CIFAR-100 has 10 times more classes than CIFAR-10, it is a considerably more difficult dataset. 
As seen from \Fref{fig:cifar100}, even previous works do not show much difference among themselves. At 20\% labeled data, LL shows slightly better performance than MCDAL. Overall, MCDAL shows the best performance across most sampling stages and shows a considerably higher performance at 40\% labeled data, with an accuracy of 67.16\% and standard deviation of $\pm0.28$. Also, similar to the result in CIFAR-10 dataset, the performance gap between MCDAL and the baseline methods is especially large in the later stages. Moreover, MCDAL is also shown to be advantageous when the amount of labeled data is low (15\% dataset labeled). Even with the minimal differences in some stages of active learning, MCDAL is shown to be promising by showing noticeable performance gap with the state-of-the-art methods especially in the early and late stages.

\begin{figure}[t]
\centering
  \includegraphics[width=.95\linewidth]{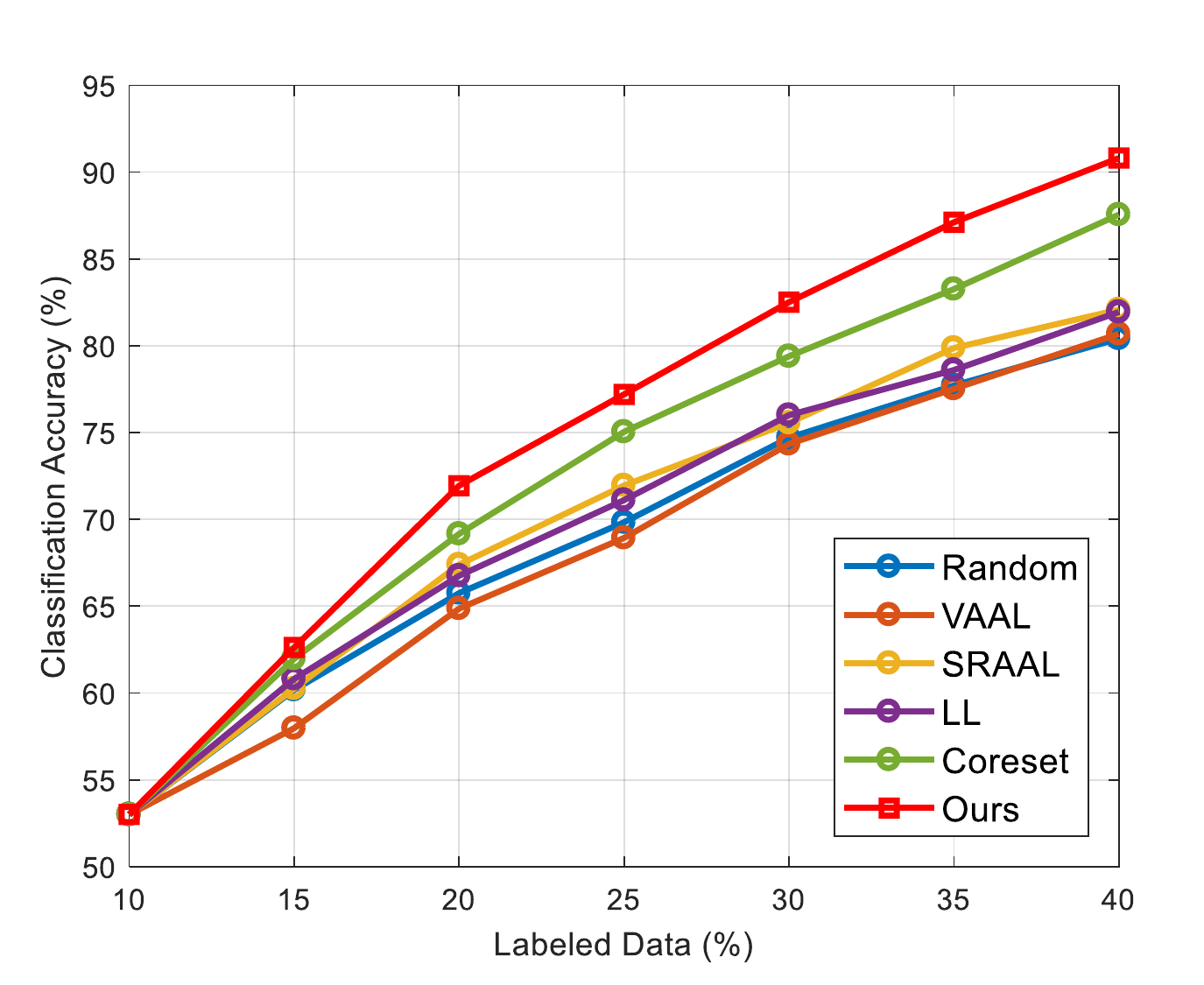}
  \caption{Comparison with existing baselines on Caltech-101. Our method, MCDAL shows the best performance compared to the state-of-the-art methods in a large margin.}
\label{fig:caltech101}
\end{figure}

\noindent\textbf{Performance on Caltech-101.}
Although the number of classes of Caltech-101 dataset is similar to that of CIFAR-100, 
the dataset is considerably smaller and each class contains a vastly different number of images.
Therefore, we believe this is an even more challenging setup and is a good testing grounds for a real-world setting where there is an imbalance of data. We can see from \Fref{fig:caltech101} that in this setup, MCDAL far outperforms previous methods at all stages and has the highest performance at 40\% labels with 90.81\% accuracy with standard deviation of $\pm0.35$.
MCDAL shows the the maximum accuracy gap of 3.84\% compared to the second-best method, Core-set approach, at 35\% labeled data.
Note that other than the Core-set approach and MCDAL show almost similar performance to that of the Random baseline.

\subsection{Semantic Segmentation}

\noindent{\textbf{Dataset.}}
For the image semantic segmentation task, we test our method on the Cityscapes~\cite{cordts2016cityscapes} dataset to compare with previous state-of-the-art works. The Cityscapes dataset consists of 3475 frames for training with instance segmentation annotations recorded in street scenes. Following the setup of prior works~\cite{sinha2019variational,zhang2020state}, we convert the annotations into 19 different classes. 

\noindent{\textbf{Comparison baselines.}}
We compare MCDAL to recent state-of-the-art methods including Query-By-Committee (QBC)~\cite{kuo2018cost}, Core-set~\cite{sener2017active}, VAAL~\cite{sinha2019variational}, and SRAAL~\cite{zhang2020state}. We also include a random sampling baseline as a standard to compare our method. To evaluate the semantic segmentation, we use mean IoU (mIoU) as the evaluation metric to measure the performance of all methods listed.

\noindent{\textbf{Implementation details.}}
We adopt the Dilated Residual Network (DRN)~\cite{yu2017dilated} as our baseline architecture.
Since we use DRN, which predicts masks for each instance, our auxiliary classification layers are the same as decoder layers of DRN. In order to apply our method, the dimension is reduced from (N,C,H,W) to (N,C) by averaging the mask. Similar to our Image Classification models, the auxiliary branches are detached so as to not affect the task model. We use Stochastic Gradient Descent as the optimizer, and set the learning rate to $1\times10^{-3}$
with a decay of $1\times10^{-1}$ for every 100 epochs. We train our method on Cityscapes for 150 epochs with a batch size 8, and we set the budget size to be 150.

\begin{figure}[t]
\centering
   \includegraphics[width=.95\linewidth]{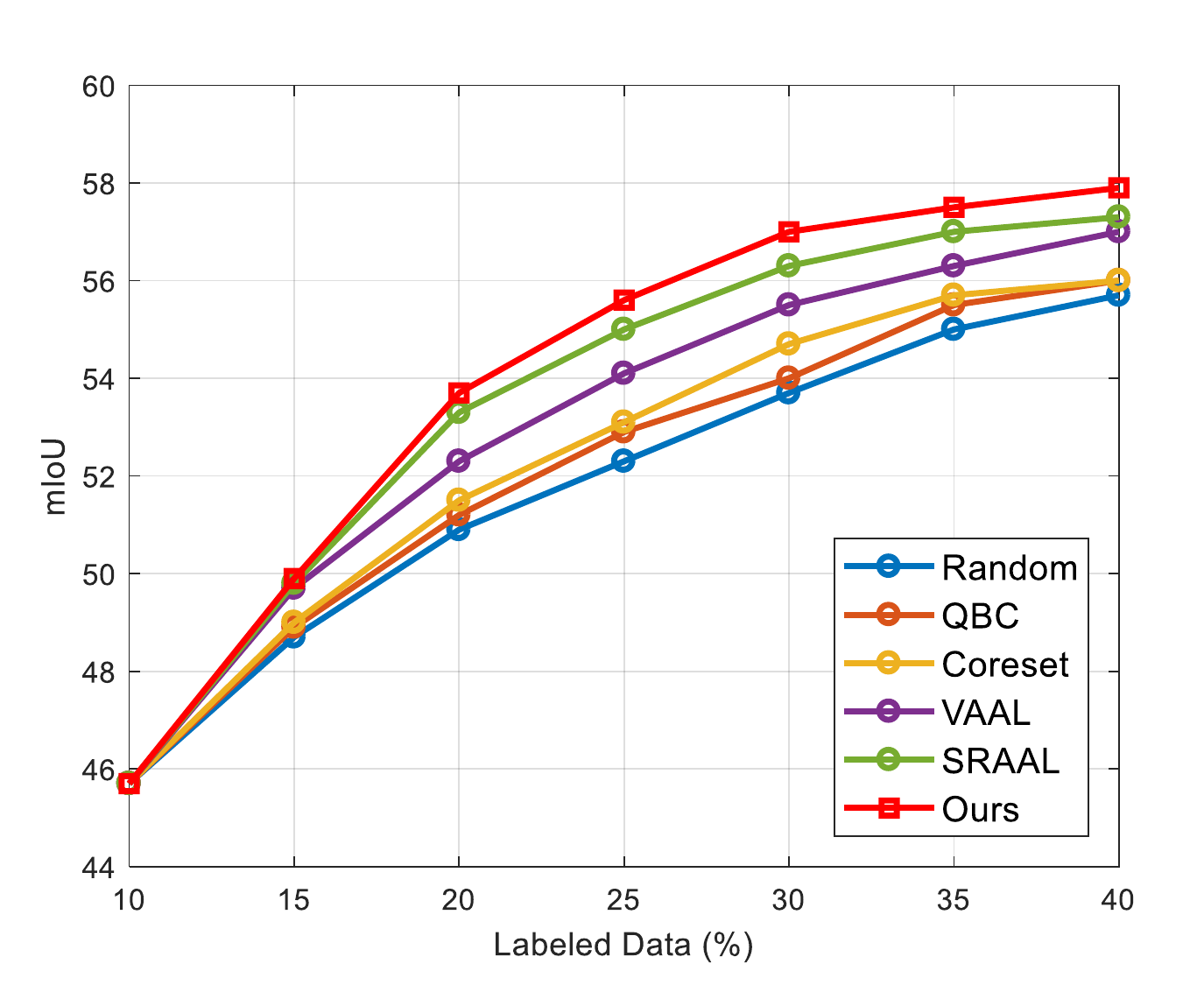}
   \caption{Comparison with existing baselines on Cityscapes dataset. MCDAL consistently shows the best performance compared to the state-of-the-art methods in a noticeable margin.}
\label{fig:cityscapes}
\end{figure}

\begin{figure*}[ht]
\centering
   \includegraphics[width=1\linewidth]{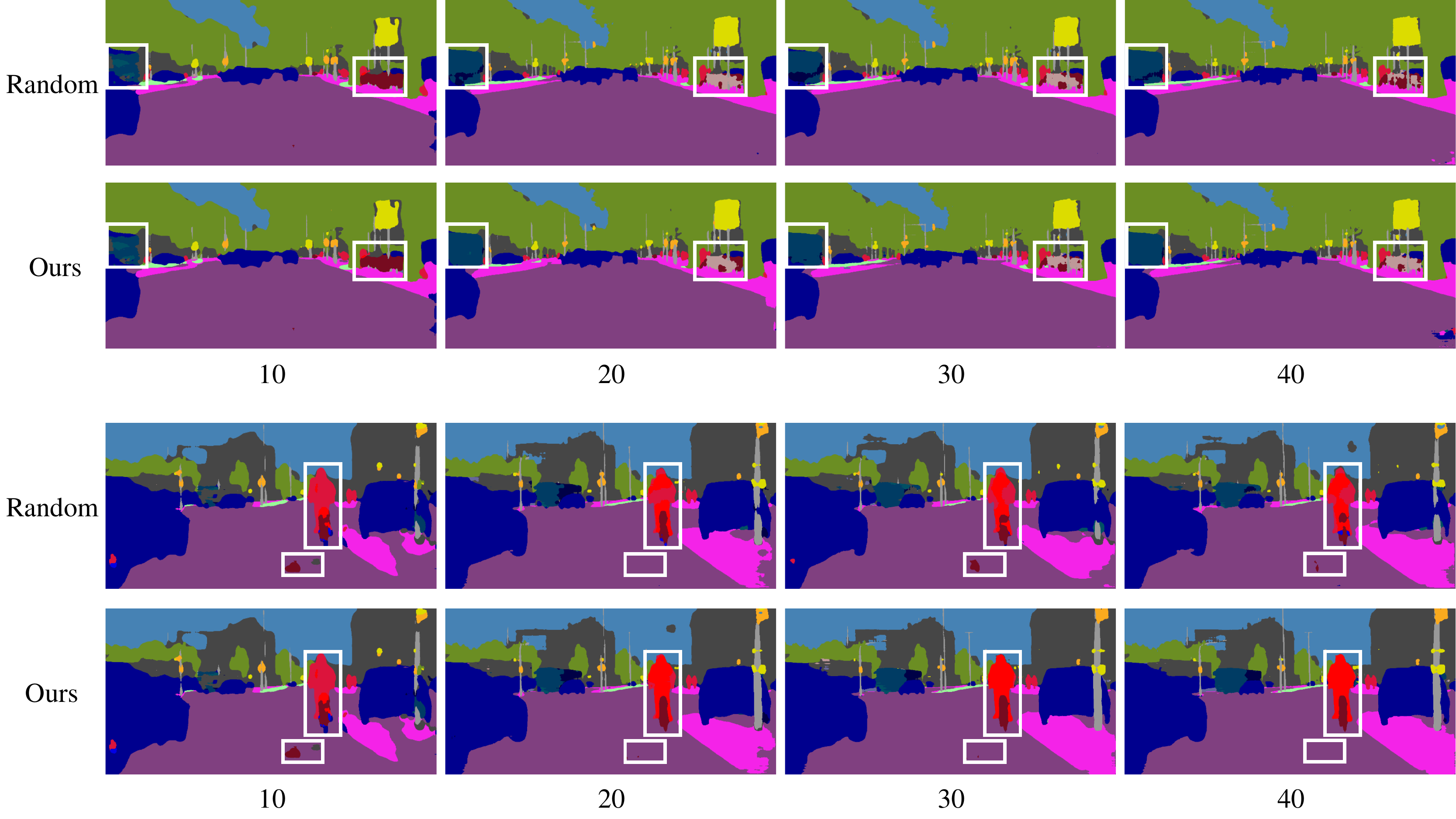}
   \caption{Qualitative results of MCDAL on Cityscapes. We show representative results in white bounding boxes to show that MCDAL is able to further refine the segmentation regions as the stages progress.
   }
\label{fig:qual}
\end{figure*}

\noindent\textbf{Segmentation performance.}
\Fref{fig:cityscapes} illustrates our results of image semantic segmentation on Cityscapes compared to previous approaches. From the figure, we can clearly see that MCDAL outperforms all other methods in all sampling stages, and MCDAL shows the highest performance at 57.9 mIoU with standard deviation of $\pm0.16$ at 40\% of the labeled data. Even though semantic segmentation is a more challenging setup compared to simple image classification, our method still shows the best performance compared to existing state-of-the-art methods by a large margin.
In addition, we also show in~\Fref{fig:qual} representative qualitative results of MCDAL in comparison to the Random baseline. We show in the white bounding boxes the regions where MCDAL is able to refine the segmentation results as the stage progresses.

\subsection{Additional Analysis}

\noindent{\textbf{Analysis on sampling time and number of parameters.}}
As active learning frameworks introduce additional sampling stages and auxiliary networks, sampling has to be fast and efficient.
In other words, one of the ultimate goals of active learning is to sample as fast as the Random baseline while being as light as possible.
\Tref{table:samplingtime} shows the sampling time required by MCDAL at \emph{stage} 1 in comparison to recent methods on CIFAR-10 datset. 
We test all methods on the CIFAR-10 dataset using a single NVIDIA TITAN Xp.
As Core-set uses an iterative method to sample, it is drastically slower than all other methods (15.8 times the sampling time required compared to MCDAL). 
VAAL and SRAAL both have VAEs and discriminators, and although faster than Core-set, is still considerably slower than our MCDAL (2.6 times and 2.8 times the sampling time required compared to MCDAL). LL shows the fastest time, which is slightly faster than ours, as LL does not compare the labeled set to the unlabeled set. 
As for number of network parameters, Core-set does not introduce any extra parameters and has the same number of parameters as a vanilla Resnet-18 network (11.17M), so we can use this as the baseline. \Tref{table:samplingtime} shows that MCDAL introduces the least number of parameters by a large margin 
even when compared to LL, while VAAL and SRAAL both show a significant increase in parameters.
In other words, we conclude that MCDAL shows the best performance with the least increase in parameters while having a reasonable sampling time.

\begin{table}[t]
\centering
    \caption{Time required to sample at \emph{stage} 1 as tested on CIFAR-10 and number of parameters of each method. MCDAL requires a reasonable sampling time. Note that GAN based state-of-the-art approaches~\cite{sinha2019variational,zhang2020state} require considerably longer sampling time in addition to a large increase in number of parameters. Note that Core-set is simply Resnet-18.}
    \vspace{2mm}
    \resizebox{1\linewidth}{!}{
    \begin{tabular}{l|| c| c}
		\hline
		Method&		Sampling Time (Sec.)&No. of Parameters\\ 
		\hline
		\hline
		Core-set~\cite{sener2017active}&114.73&11.17M\\
		LL~\cite{yoo2019learning}&6.53&11.29M\\
		VAAL~\cite{sinha2019variational}&18.94&34.29M\\
		SRAAL~\cite{zhang2020state}&20.48&45.58M\\
		\hline
		\textbf{MCDAL (Ours)}&7.25&11.19M\\
		\hline
	\end{tabular}
	}
\label{table:samplingtime}
\end{table}%



\begin{figure}[t]
\centering
   \includegraphics[width=.95\linewidth]{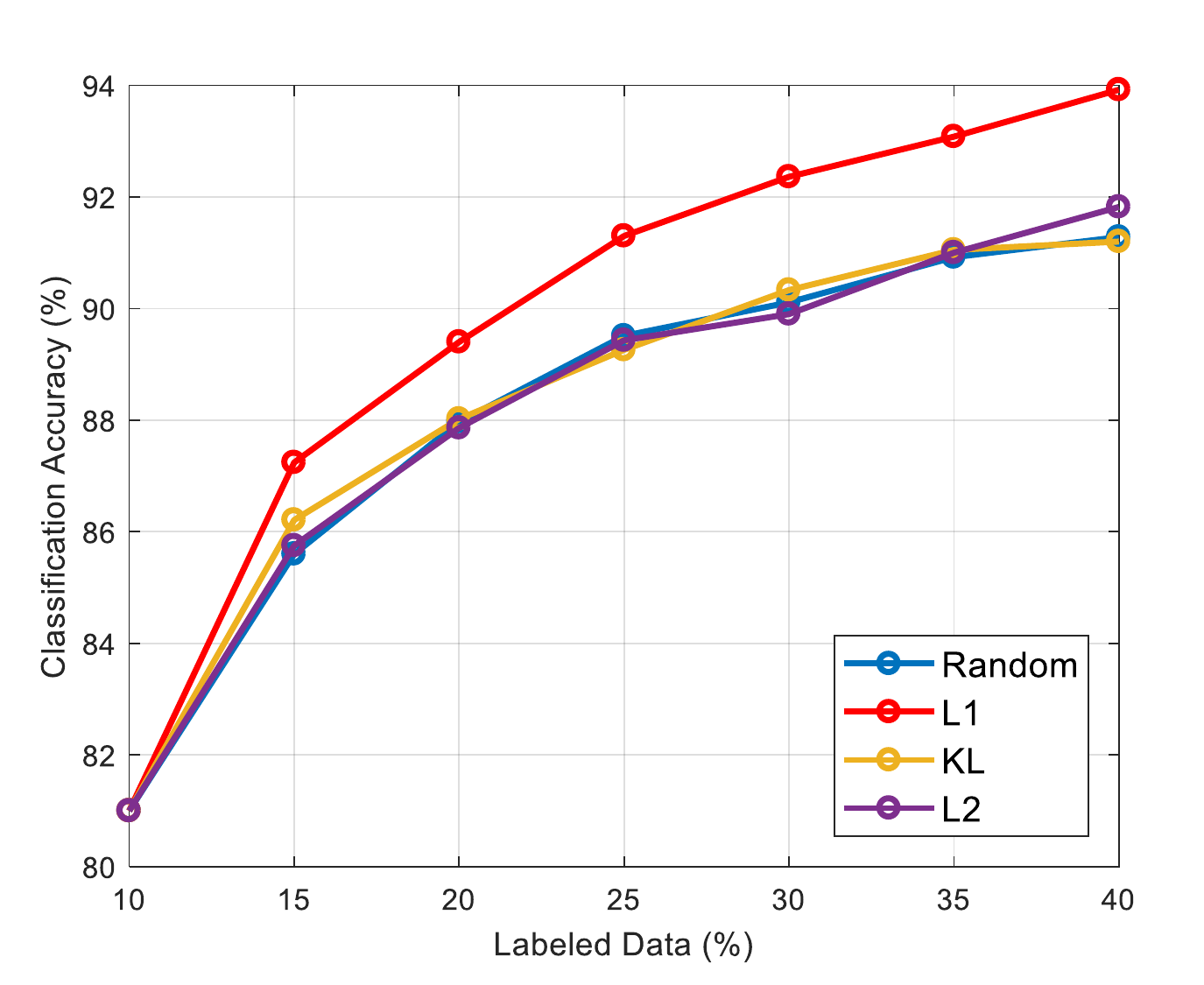}
   \caption{Results of the variants of our method with different distance measure for $d(\cdot,\cdot)$ (L1, L2, KL-divergence). L2 and KL-divergence show similar performance as the Random baseline, showing that it cannot be considered as a design choice.}
\label{fig:distance}
\end{figure}

\noindent\textbf{Comparison on different distance measure.}
Based on the theory by~\cite{ben2010theory}, we chose our discrepancy measure for $d(\cdot,\cdot)$ to be the L1 distance as in \Eref{eqn:l1dist}; however, we perform an ablation study on CIFAR-10 with other discrepancy measures such as KL-divergence (KL), which is broadly used in knowledge distillation~\cite{cho2021dealing,hinton2015distilling,kim2018disjoint,kim2020detecting,kim2021acp++}, and L2 distance with our final design choice of L1. 
\Fref{fig:distance} shows that KL and L2 perform poorly and shows a similar performance to the Random baseline, showing that using these distance measures is meaningless as the Random baseline is much simpler and faster in every way. 
Empirically, we show that L1 is the only viable design choice among the three options evaluated.

\begin{figure}[t]
\centering
   \includegraphics[width=.95\linewidth]{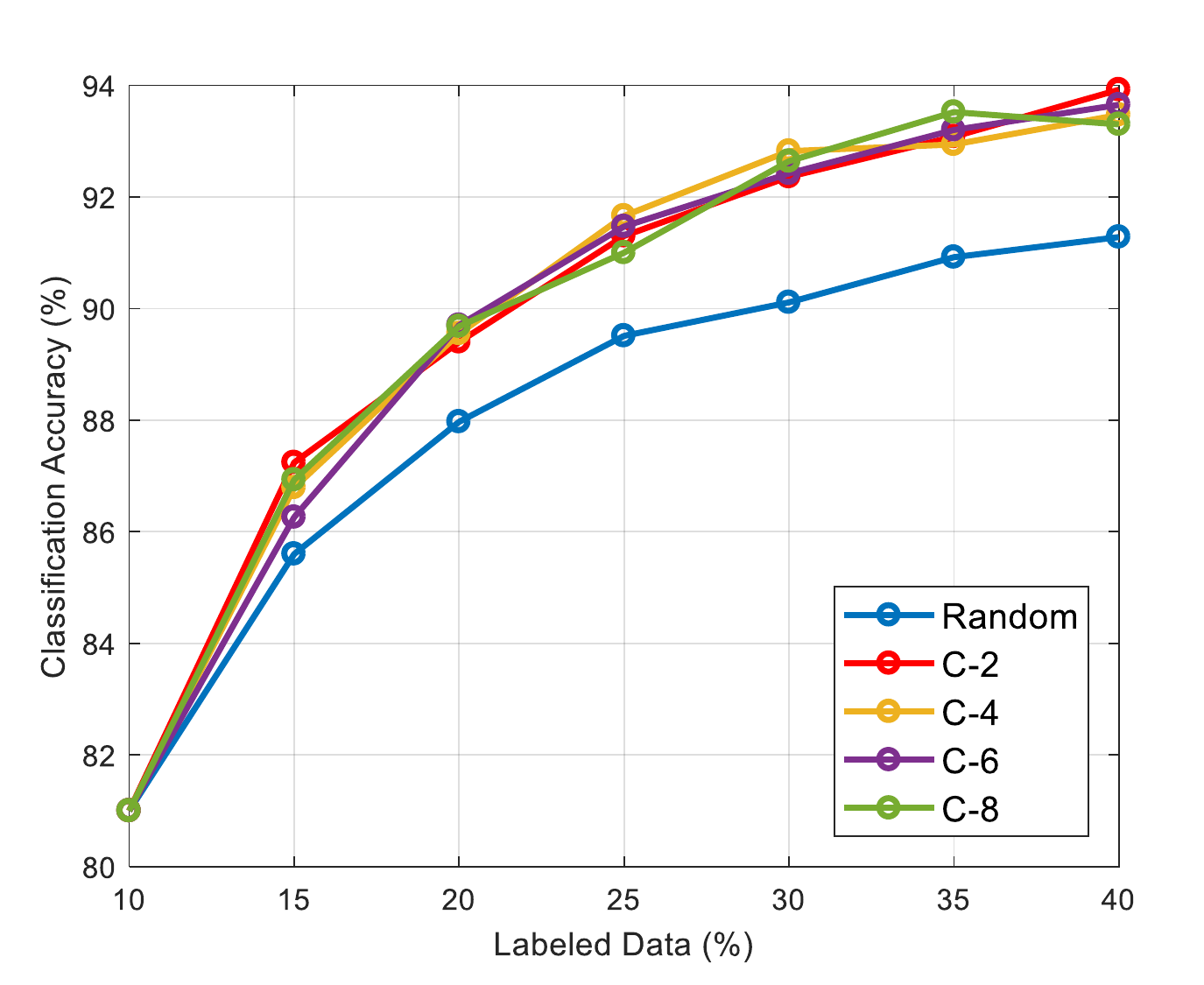}
   \caption{
            Results of our method with varying number of the auxiliary classification layers, two ($F_{1\sim 2}$), four ($F_{1\sim 4}$), six ($F_{1\sim 6}$), and eight ($F_{1\sim 8}$) show similar performances.
            In light of this finding, for the other experiments,
            we select the number of the auxiliary layers to be \emph{two} due to the model complexity.}
\label{fig:num_of_cls}
\end{figure}

\noindent\textbf{Comparison on number of classifiers.}
In addition, we perform an ablation study to see how changing the number of auxiliary layers set $|F_{\{\cdot\}}|$ would affect the active learning performance of our method. As our auxiliary classification layers are in pairs, we test the different number of the auxiliary layers in pairs to see the affect on the performance, two ($F_{1\sim 2}=\{F_1,F_2\}$), four ($F_{1\sim 4}=\{F_1,F_2,...,F_4\}$), six ($F_{1\sim 6}=\{F_1,F_2,...,F_6\}$), and eight ($F_{1\sim 8}=\{F_1,F_2,...,F_8\}$). We test this ablation on the CIFAR-10 dataset and the result is illustrated in \Fref{fig:num_of_cls}.
\Fref{fig:num_of_cls} shows that increasing the number of classification layers has little to no effect on performance. 
Regardless of the number of auxiliary layers we use, the performance is similarly favorable compared to other active learning methods.
This shows that our method already functions favorably with 2 classifiers,
and we conjecture that using more classifiers seems to introduce a form of redundancy and not introduce any new information in regards to finding more informative samples,
therefore, we set the number of auxiliary classification layers to two ($\{F_1, F_2\}$), due to the time complexity and memory cost of training.


\begin{figure}[t]
\centering
   \includegraphics[width=.95\linewidth]{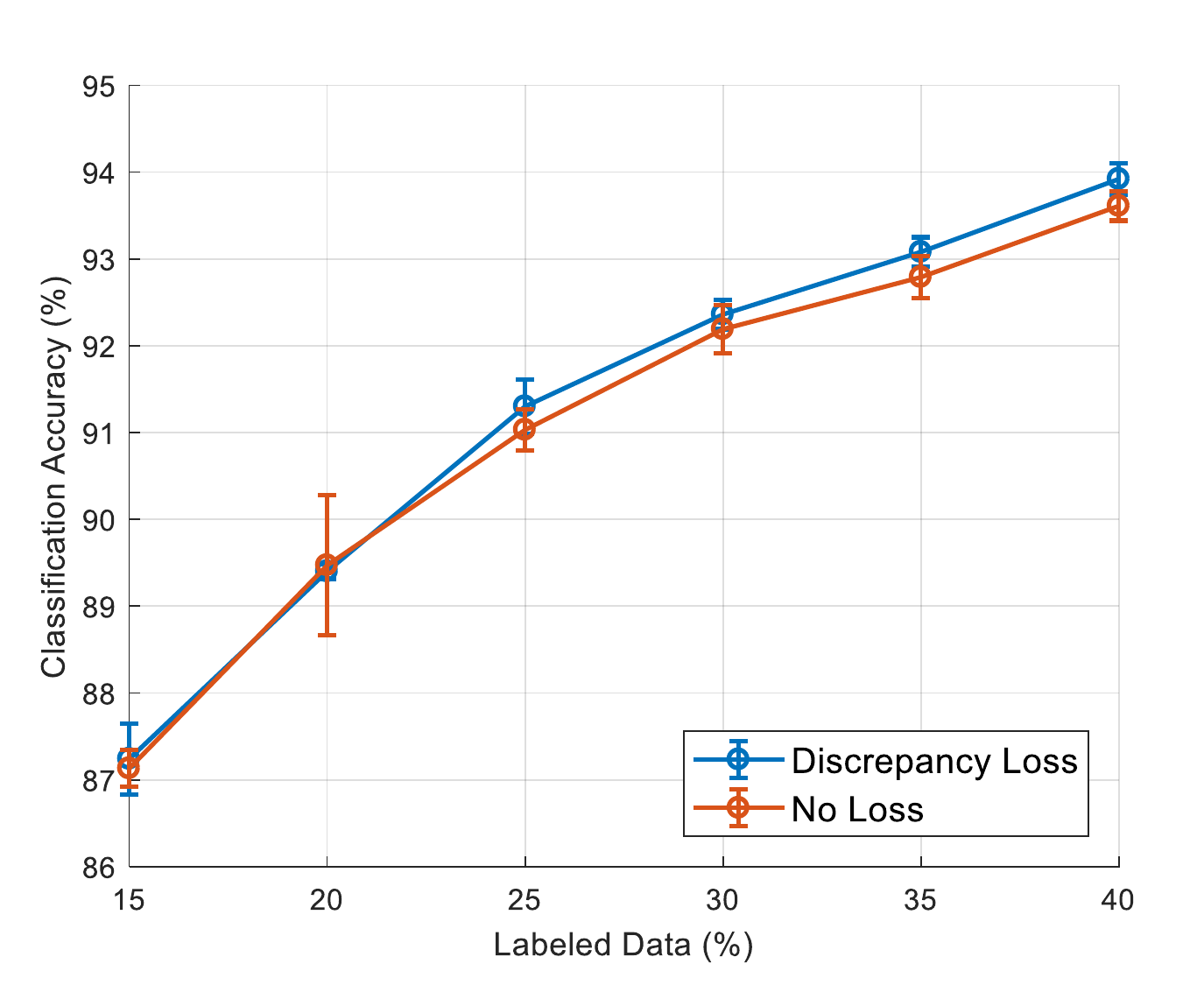}
   \caption{Results of ablation of with and without discrepancy loss, as the first stage is the same and to show the standard deviation of each stage more clearly, we show only from 15\% to 40\%. Our method without the discrepancy loss shows similar average performance, but the addition of the discrepancy loss aids heavily in the stability of MCDAL.}
\label{fig:lossabla}
\end{figure}

\noindent\textbf{Ablation of the discrepancy loss.}
To understand the effects of the discrepancy loss, we additionally perform an ablation of MCDAL with and without the discrepancy losses and show our findings in \Fref{fig:lossabla}. For better visualization of the standard deviation error bars and as the first stage is fixed, we start the graph at 15\% labeled data. We find that the average performance of our method without the discrepancy loss is similar to that of ours with the loss. However, including the loss significantly increases stability especially in the earlier stages as shown in the figure. Hence, we propose that the discrepancy loss is a helpful addition.

\label{sec.experiments}

\section{Conclusion}
In this paper, we introduce a novel active leaning framework, Maximum Classifier Discrepancy for Active Learning (MCDAL), that does not rely on Generative Adversarial Networks (GAN) and still outperforms the recent state-of-the-art models with lesser parameters, easy implementation, and fast sampling.
We show in our experimental results that our model 
is easily and readily applicable to any and all different kinds of datasets and tasks. We show through our findings that although GAN methods may seem like the way forward, our method shows promise and might be an alternative for future active learning research.

\noindent\textbf{Acknowledgements} 
This work was supported in part by the Institute of Information and Communications Technology Planning and Evaluation (IITP) Grant funded by the Korea Government (MSIT) (Artificial Intelligence Innovation Hub) under Grant 2021-0-02068, and also under the framework of international cooperation program managed by the National Research Foundation of Korea (NRF2020M3H8A1115028, FY2021).

\ifCLASSOPTIONcaptionsoff
  \newpage
\fi



%


{
\bibliographystyle{IEEEtrans}
\bibliography{egbib}
}
%

\begin{IEEEbiography}[{\includegraphics[width=1in,height=1.25in,clip,keepaspectratio]{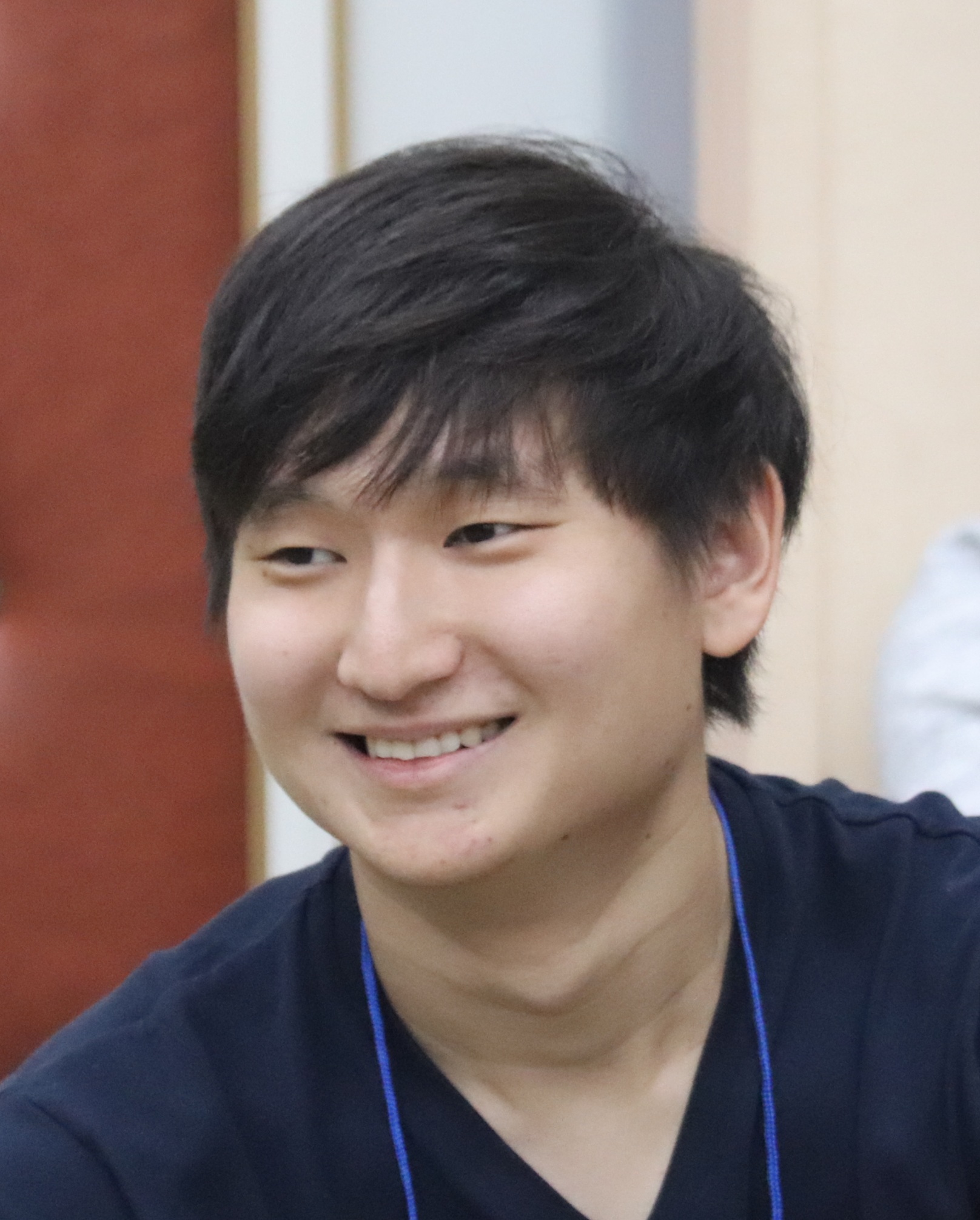}}]{Jae Won Cho}
received the B.S. degree in Electrical Engineering from Georgia Institute of Technology, Atlanta, GA, USA in 2018,. He is currently a Ph.D. candidate in Electrical Engineering at the Korea Advanced Institute of Science and Technology (KAIST) under the supervision of Professor In So Kweon. He was awarded a bronze prize from Samsung Humantech paper awards. His current research interests include deep learning topics such as active learning and high-level computer vision application such as vision and language. He is a student member of the IEEE.
\end{IEEEbiography}

\begin{IEEEbiography}[{\includegraphics[width=1in,height=1.25in,clip,keepaspectratio]{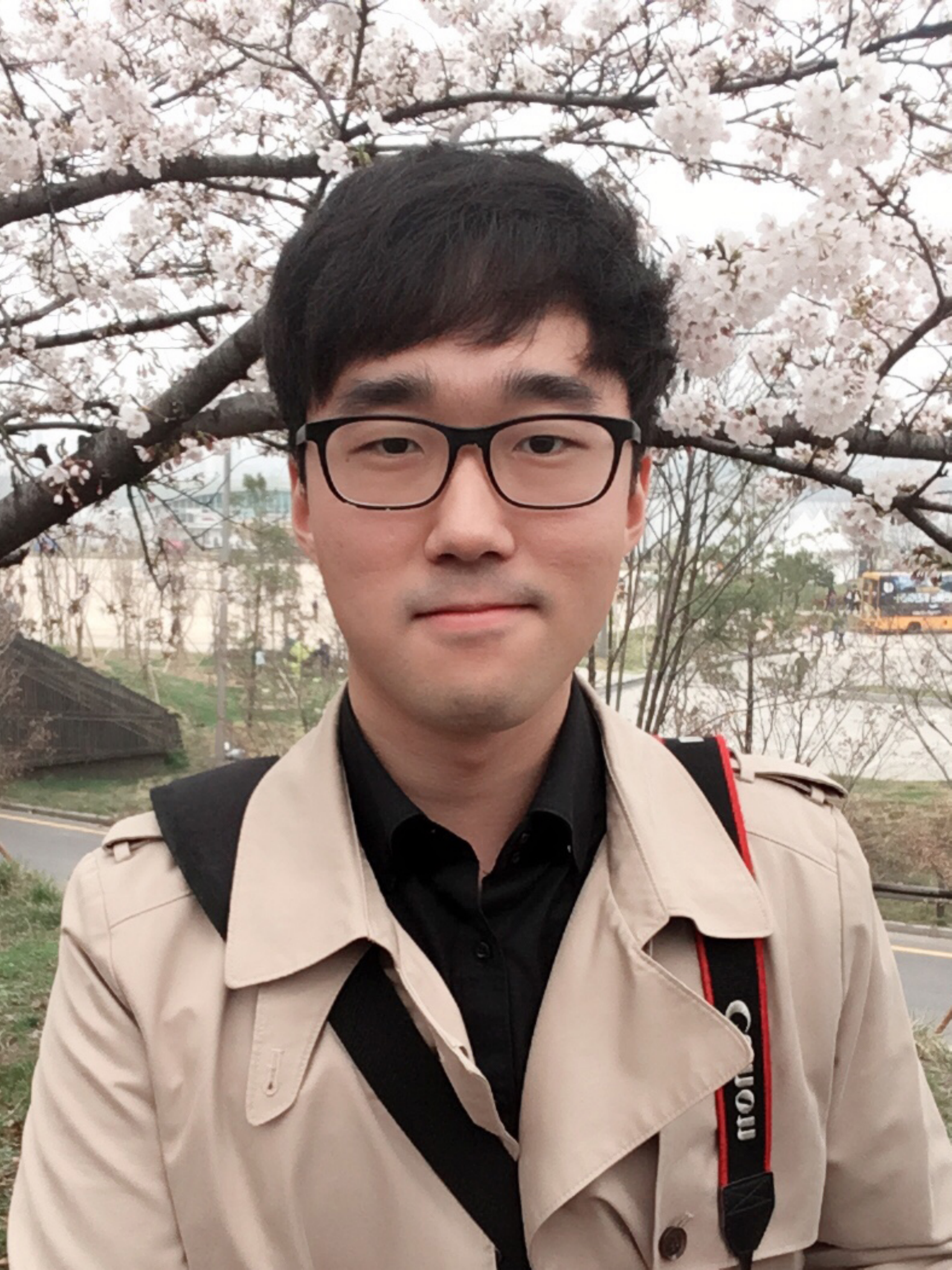}}]{Dong-Jin Kim}
received the B.S. degree, M.S. degree, and Ph.D. degree in Electrical Engineering from Korea Advanced Institute of Science and Technology (KAIST), Daejeon, South Korea, in 2015, 2017, and 2021, respectively.
He is currently a Postdoc in EECS at UC Berkeley.
He was a research intern in the Visual Computing Group, Microsoft Research Asia (MSRA).
He was awarded a silver prize from Samsung Humantech paper awards and Qualcomm Innovation awards.
His research interests include data issues in computer vision especially in high-level computer vision problems.
He is a member of the IEEE.
\end{IEEEbiography}


\begin{IEEEbiography}[{\includegraphics[width=1in,height=1.25in,clip,keepaspectratio]{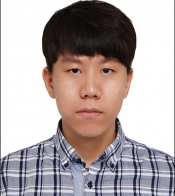}}]{Yunjae Jung}
received the B.S. degree in Electrical Engineering from Sogang University, Seoul, South Korea in 2018 and reeived M.S. degree in Electrical Engineering from Korea Advanced Institute of Science and Technology (KAIST), Daejeon, South Korea, in 2019. He is currently working towards the Ph.D. degree in Electrical Engineering at KAIST. He was awarded a honorable mention from Samsung Humantech paper awards. His research interests include high-level computer vision such as video scene understanding and vision and language. He is a student member of the IEEE.
\end{IEEEbiography}

\begin{IEEEbiography}[{\includegraphics[width=1in,height=1.25in,clip,keepaspectratio]{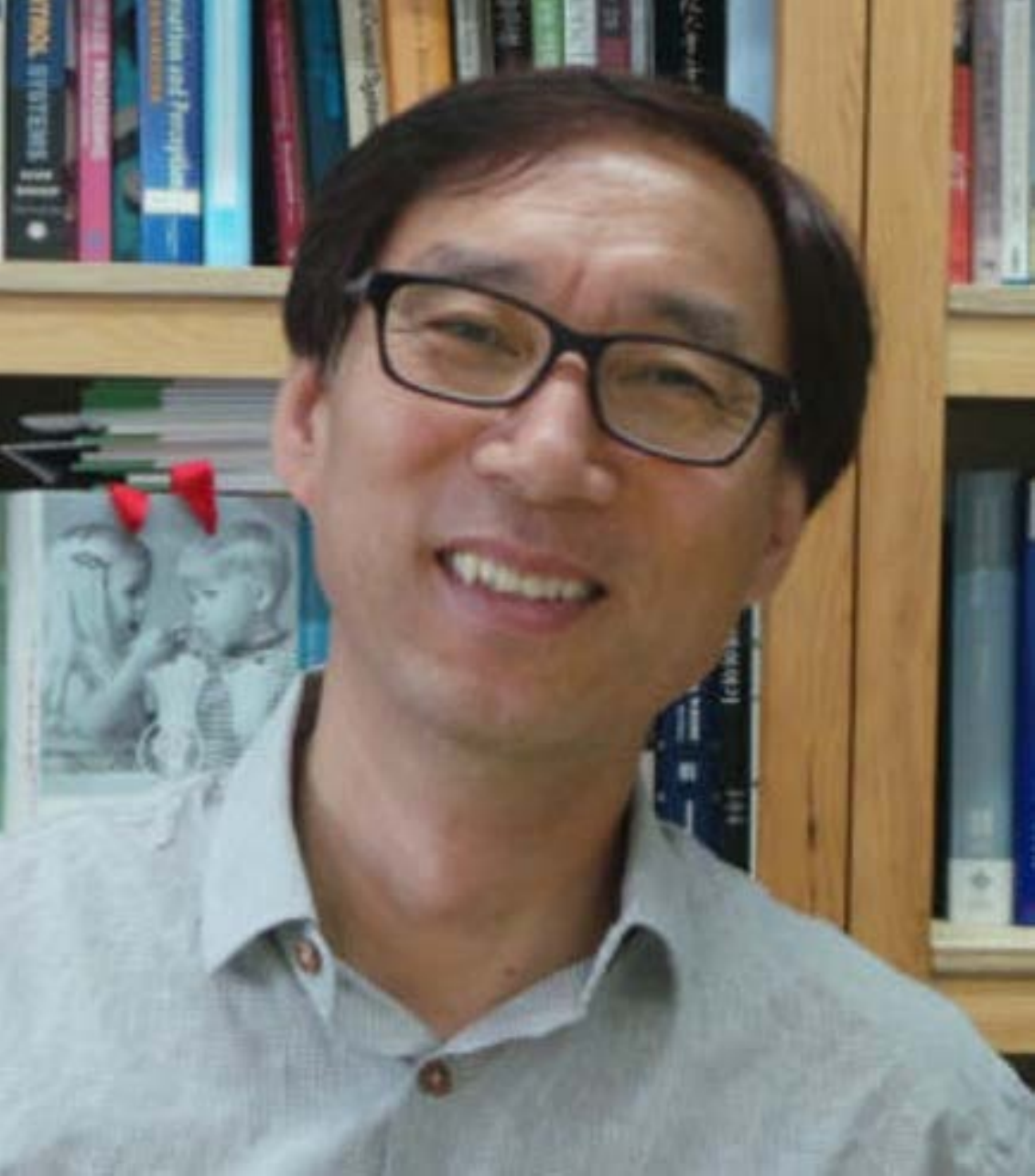}}]{In So Kweon}
received the B.S. and M.S. degrees in mechanical design and production engineering from Seoul National University, South Korea, in 1981 and 1983, respectively, and the Ph.D. degree in robotics from the Robotics Institute, Carnegie Mellon University, USA, in 1990.
	He was with the Toshiba R\&D Center, Japan, and he is currently a KEPCO chair professor with the Department of Electrical Engineering, since 1992. 
	He served as a program co-chair for ACCV 07' and ICCV 19', and a general chair for ACCV 12'. 
	He is on the honorary board of IJCV. 
	He was a member of ``Team KAIST,'' which won the first place in DARPA Robotics Challenge Finals 2015. 
	He is a member of the IEEE and the KROS.
\end{IEEEbiography}



\end{document}